\documentclass[12pt]{article}

\usepackage[utf8]{inputenc}
\usepackage[british]{babel}
\usepackage{amsmath, amsthm, amssymb}
\usepackage{graphicx}
\usepackage[hidelinks]{hyperref}
\usepackage{setspace}
\usepackage{geometry}
\usepackage{amsmath}
\usepackage{amssymb}
\usepackage{amsthm}
\usepackage[hidelinks]{hyperref}
\usepackage{float}

\newtheorem{proposition}{Proposition}
\usepackage{amsthm}

\title{Authority-Level Priors: An Under-Specified Constraint in Hierarchical Predictive Processing}

\author{
    Marcy Palejova\\
    \small Anglia Ruskin University\\
    \small \texttt{mmp144@pgr.aru.ac.uk}
}

\date{\today}

\begin{document}

\maketitle

\begin{abstract}
\noindent

Hierarchical predictive processing explains adaptive behaviour through precision-weighted inference within a single generative model. Yet explicit belief revision and increased confidence in adaptive self-models often fail to produce corresponding changes in stress reactivity or autonomic regulation. This asymmetry suggests that the framework leaves formally under-specified a governance-level constraint concerning which identity-level hypotheses are permitted to regulate autonomic and behavioural control under uncertainty.

We introduce Authority-Level Priors (ALPs) as meta-structural constraints defining a regulatory-admissible subset $\mathcal{H}_{\text{auth}} \subset \mathcal{H}$ of identity-level hypotheses. ALPs are not additional representational states, nor hyperpriors over precision; rather, they constrain which hypotheses are admissible for regulatory control, independent of precision-weighted inference operating within that admissible set. Precision determines influence conditional on admissibility, whereas ALPs determine admissibility itself.

This distinction explains why explicit belief updating and repeated practice readily modify representational beliefs while autonomic threat responses and identity-level predictions often remain stable despite contradictory evidence and sustained effort. We provide a minimal computational formalisation in which policy optimisation is restricted to policies generated by authorised hypotheses, yielding testable predictions concerning stress-reactivity dynamics, recovery time constants, compensatory control engagement, and cross-context behavioural persistence.

Neurobiologically, ALPs are hypothesised to manifest functionally through distributed prefrontal arbitration and control networks involved in rule maintenance, conflict monitoring, and top-down modulation. The proposal is compatible with variational active inference and introduces no additional inferential operators, instead formalising a boundary condition required for determinate identity–regulation mapping when multiple high-precision hypotheses coexist.

The model generates falsifiable predictions: governance shifts should produce measurable changes in stress-reactivity curves, recovery dynamics, compensatory cognitive effort, and the durability of behavioural change under identical load conditions. ALPs are advanced as an architectural hypothesis to be evaluated through computational modelling and longitudinal stress-induction paradigms.
\end{abstract}

\vspace{0.5cm}

\vspace{1cm}

\section{Background and the Identified Gap}

\subsection{Predictive Processing and Identity-Level Regulation}

The present analysis restricts scope to systems with identity-level self-models - specifically humans, and potentially non-human primates where analogous regulatory structures may exist. Hierarchical predictive processing frameworks model perception, action, and learning as inference under the Free Energy Principle, where the brain minimizes prediction error through precision-weighted belief updating across nested temporal and spatial scales (Friston, 2010; Clark, 2016; Hohwy, 2013). Higher levels of the cortical hierarchy generate predictions about lower-level activity, while ascending prediction errors signal departures from expectation, with precision determining the relative influence of top-down priors versus bottom-up sensory signals (Feldman \& Friston, 2010). Active inference extends this account to action selection, proposing that agents minimize expected free energy by sampling the environment in ways that reduce both uncertainty (epistemic value) and distance from preferred states (pragmatic value; Friston et al., 2017; Parr \& Friston, 2019). Control hierarchies emerge naturally within this architecture, with higher-order policies modulating lower-level reflexes and habits through context-dependent precision adjustments (Pezzulo et al., 2018). This framework offers a principled, neurobiologically grounded account of cognitive control, affective regulation, and adaptive behavior in humans. 

It explains resistance to belief updating through entrenched high-precision priors that down-weight prediction errors inconsistent with the existing model, accounts for the persistence of maladaptive predictions via learned precision assignments, and models therapeutic change as gradual recalibration of hierarchical generative models through accumulated evidence (Barrett, 2017; Seth \& Friston, 2016).
Yet despite its explanatory breadth, predictive processing leaves implicit - and therefore formally under-specified - a governance-level constraint that determines which identity-level hypotheses within the generative model are authorized to regulate identity, affect, and autonomic systems when multiple priors simultaneously compete for behavioral control in human agents.
Consider the clinical asymmetry frequently observed in therapeutic contexts: clients can acquire explicit insight into maladaptive patterns, endorse corrective beliefs through conscious reflection, and demonstrate comprehension of alternative self-models, yet autonomic threat responses, affective predictions, and identity-level regulatory patterns persist unchanged despite sustained contradictory evidence and deliberate cognitive effort (Hayes et al., 2011; Bouton, 2002). Standard predictive processing accounts attribute this persistence to high-precision trauma priors or context-dependent precision reweighting—explanations that capture what remains stable but do not specify why certain priors maintain regulatory dominance even when alternative hypotheses accumulate evidence and precision within representational domains, nor why sudden, durable stabilization occasionally occurs without the gradual evidence accumulation that Bayesian updating typically requires.
The framework provides no formal mechanism that distinguishes which predictions can stabilize regulatory systems from which predictions are merely represented, endorsed, or consciously believed. A human agent may hold multiple contradictory identity-level priors simultaneously - "I am competent" (consciously endorsed) and "I am inadequate" (autonomically predicted) - yet one dominates regulatory control over stress reactivity, affective tone, and behavioral selection while the other remains inert despite equal or greater explicit conviction. Current models attribute this to precision differences, but precision modulates influence within an already-authorized generative model; it does not explain which models are candidates for precision assignment in the first place, nor what determines regulatory dominance when precision values are comparable.

\subsection{Governance-Level Under-Specification}

Three structural incompletenesses emerge at the governance level of hierarchical predictive processing:

First, the framework does not formally define what constrains the evaluable policy space during active inference. Expected free energy calculations assume a pre-defined set of candidate policies (Friston et al., 2015), but the architecture provides no principled account of which policies enter this set under identity-threatening uncertainty versus which remain outside consideration regardless of their computed value. When a human faces situations that challenge core self-models - relational rejection, competence threats, existential uncertainty - certain behavioral policies become subjectively "unthinkable" or "impossible" despite lower expected free energy than enacted alternatives. This is not merely high prediction error; it reflects architectural exclusion from the hypothesis space within which optimization occurs—policies remain excluded even when they would minimise expected free energy. Current models assume policy priors but do not specify what authorizes certain priors to regulate behavior while excluding others from evaluation entirely.
Second, the framework does not account for immediate, durable stabilization without gradual Bayesian updating. Clinical observations document instances where identity-level predictions shift rapidly and persist across contexts without reinforcement, accompanied by measurable reductions in autonomic reactivity and compensatory cognitive effort (Wilkinson-Ryan \& Westen, 2000; sudden gains literature: Tang \& DeRubeis, 1999). Standard explanations invoke extreme precision reassignment or complete model replacement, yet neither mechanism explains how such transitions occur without catastrophic disruption to ongoing prediction or why they produce greater stability than gradual updating.
If precision reweighting alone were sufficient, identical outcomes should follow from sustained cognitive effort - yet clinically, effort-based interventions often fail to produce comparable durability, collapsing under stress despite maintained conscious conviction.
Third, the architecture does not differentiate regulatory authority from representational content. A human agent's self-model may include multiple identity priors - some consciously endorsed, some autonomically enacted, some historically accurate, some aspirational. Predictive processing explains how these coexist through precision hierarchies and contextual gating, but it does not specify what determines which prior governs autonomic, affective, and behavioral systems when conscious endorsement and autonomic prediction conflict. Metacognitive monitoring can detect this mismatch (Barrett \& Simmons, 2015), yet detection does not resolve the governance conflict. Insight into the discrepancy - "I know I am safe, yet my body responds as threatened" - leaves the regulatory structure unchanged. Something beyond precision weighting, hierarchical level, and metacognitive awareness determines permission to regulate, distinct from permission to represent. 
These three gaps converge on a single under-specification: hierarchical predictive processing does not formally define the highest-order constraint that determines which generative models are authorized to stabilize identity, affect, and autonomic regulation under conditions where multiple high-level priors compete for control. We propose Authority-Level Priors (ALPs) as the formalization of this governance layer - not as an addition to the hierarchy, but as the explicit definition of a constraint the architecture implicitly requires yet leaves architecturally unspecified.

\section{Definition of Authority-Level Priors}

We define Authority-Level Priors (ALPs) as meta-structural constraints within hierarchical predictive processing that determine which high-level identity hypotheses are permitted to exert regulatory control over autonomic, affective, and behavioural systems, particularly under conditions of identity-relevant uncertainty. ALPs operate at a level architecturally orthogonal to posterior probability and precision-weighting dynamics, constraining regulatory admissibility rather than modulating inferential confidence within an already-authorized hypothesis space.
Within the predictive processing framework, a human agent is modelled as maintaining a single hierarchical generative model in which multiple internally coherent high-level identity hypotheses can coexist. These hypotheses generate distinct predictions about self-capability, relational safety, competence under stress, and worthiness of social inclusion. Standard accounts assume such hypotheses compete through Bayesian belief updating, with posterior probability and precision determining relative influence over prediction and action. However, clinical observations reveal a persistent asymmetry; multiple identity hypotheses can coexist with comparable posterior probability and representational precision, yet differ categorically in their capacity to regulate autonomic systems, affective tone, and behavioural policy selection under stress. One hypothesis dominates regulatory control while others remain representationally active but behaviourally inert, despite equal or greater conscious endorsement and evidential support.
ALPs formalize the governance layer that resolves this asymmetry. Formally, let $\mathcal{H}$ represent the set of all coherent identity-level hypotheses within the generative model, and let $\mathcal{H}_{\text{auth}} \subset \mathcal{H}$ denote the subset of hypotheses authorized to regulate autonomic and behavioural systems at time $t$. ALPs define the constraint structure determining membership in $\mathcal{H}_{\text{auth}}$, without altering the inferential updating of hypotheses within $\mathcal{H}$. A hypothesis $h_i \in \mathcal{H}$ may carry high posterior probability $P(h_i \mid \text{data})$ and high representational precision $\pi_i$, yet if $h_i \notin \mathcal{H}_{\text{auth}}$, it remains excluded from regulatory influence regardless of its inferential strength. Conversely, a hypothesis $h_j \in \mathcal{H}_{\text{auth}}$ exerts regulatory control even when alternative hypotheses exhibit higher posterior probability within representational domains.
This distinction is categorical, not continuous. ALPs do not modulate how strongly a hypothesis influences regulation (a function of precision); they determine whether a hypothesis is a candidate for regulatory influence. The constraint operates at the level of admissibility structure rather than inferential dynamics, functioning as a boundary condition on regulatory admissibility rather than as an inferential operator within the generative model. Critically, ALPs preserve Bayesian updating over hypothesis content---belief revision remains continuous and evidence-driven for all hypotheses in $\mathcal{H}$. What changes discontinuously during governance shifts is not inferential dynamics but the regulatory permission structure: which hypotheses transition into or out of $\mathcal{H}_{\text{auth}}$---analogous to discrete attractor-basin transitions in dynamical systems.
At any given moment, regulatory stabilization at time $t$ reflects dominance by a singular governance regime. While multiple identity hypotheses may remain representationally active and contextually modulated, regulatory coherence across autonomic set-points, affective predictions, and stress-response trajectories is maintained through singular governance dominance, reducing destabilizing oscillation between incompatible physiological states. The governance regime may shift---sometimes rapidly and durably---but the regulatory state at any point reflects control by a hypothesis within $\mathcal{H}_{\text{auth}}$.

To clarify scope, Authority-Level Priors are not proposed as additional representational states within the generative model, nor as hyperpriors over precision or policy priors within active inference. Rather, ALPs formalize a constraint on regulatory admissibility that determines which identity-level hypotheses are eligible to interface with autonomic and behavioural control systems. Precision-weighted inference, policy optimisation, and contextual modulation continue to operate within this admissible hypothesis set. The proposal therefore does not modify the inferential calculus of predictive processing; it specifies a boundary condition governing which hypotheses participate in regulatory control when multiple high-level identity hypotheses coexist.

\begin{proposition}
Precision determines influence \emph{within} the regulatory-admissible hypothesis set, whereas Authority-Level Priors (ALPs) determine which hypotheses constitute that set.
\end{proposition}


Why ALPs Are Architecturally Necessary

The necessity of governance-level constraints emerges from the functional requirements of regulatory coherence in systems with identity-level self-models. Precision weighting alone does not formally specify how cross-system regulatory dominance is determined when multiple high-level identity hypotheses maintain comparable inferential strength. Consider a human agent maintaining two coherent hypotheses: $h_{\text{safe}}$ (``I am relationally secure'') and $h_{\text{threat}}$ (``I am socially vulnerable''). Both may be contextually appropriate, evidentially supported, and assigned high precision in their respective domains. Standard predictive processing frameworks imply that precision weighting will determine regulatory influence in practice---whichever hypothesis carries higher precision should dominate autonomic set-points and affective tone.

However, clinical observations reveal this mechanism is under-specified for identity-level regulation. A client may consciously endorse $h_{\text{safe}}$ with high precision (strong conviction, low uncertainty, substantial therapeutic evidence), yet autonomic systems continue operating under $h_{\text{threat}}$ with equal precision (chronic hypervigilance, elevated baseline cortisol, stress-related autonomic activation consistent with limbic threat-processing engagement).Both hypotheses remain high-precision; both generate coherent predictions; neither is eliminated through evidence accumulation. If precision weighting were sufficient, these hypotheses should alternate influence based on contextual fluctuations - yet regulatory dominance persists across contexts without continuous environmental reinforcement, demonstrating that a governance-level constraint beyond precision weighting determines which hypothesis governs autonomic regulation.

In the absence of an explicit regulatory-admissibility constraint, the architecture is underdetermined at the identity-regulation interface: when multiple high-precision identity hypotheses are simultaneously active, the framework specifies how precision weights prediction errors within a hypothesis but does not specify which hypotheses are eligible to set autonomic and affective set-points. ALPs formalise this missing admissibility constraint, yielding a determinate mapping from identity-level inference to regulatory control that the architecture otherwise leaves unspecified. Systems that regulate across multiple physiological and behavioural subsystems require a stable control reference to coordinate prediction across timescales. ALPs are proposed as the formal specification of that control reference at the identity level.
ALPs resolve this under-specification by constraining which identity-level hypotheses are eligible to interface with autonomic and behavioural control systems. Precision determines influence conditional on admissibility, whereas Authority-Level Priors determine admissibility itself. Precision modulates confidence within an authorized hypothesis set; ALPs determine membership in that set. This architectural constraint provides the stability necessary for coherent cross-system regulation when multiple high-precision identity hypotheses coexist.

ALPs vs. Precision: The Critical Distinction
Precision and ALPs perform categorically distinct functions within predictive processing architecture:
Precision specifies the confidence weighting assigned to prediction errors generated by a hypothesis within an authorized hypothesis space. High precision increases the influence of prediction errors generated by that hypothesis, driving belief updating and policy selection within its domain. Precision is a scalar weighting parameter that modulates influence continuously and can be adjusted through evidence accumulation or contextual shifts (Feldman \& Friston, 2010).
ALPs specify which hypotheses are authorized to regulate autonomic and behavioural systems. This is a categorical constraint determining set membership in $\mathcal{H}_{\text{auth}}$, not a continuous weighting function. Two hypotheses may carry identical precision values yet differ in regulatory authority: one generates predictions that modulate autonomic tone, affective set-points, and stress reactivity; the other generates predictions that remain representationally active but do not interface with regulatory systems.
The empirical signature of this distinction appears when conscious belief updates without autonomic stabilization. A human agent can increase precision on $h_{\text{safe}}$ through therapeutic work---reducing uncertainty, accumulating confirming evidence, strengthening conviction---yet if $h_{\text{safe}} \notin \mathcal{H}_{\text{auth}}$, this precision increase does not translate to regulatory control. The hypothesis influences conscious reflection and explicit reasoning (representational domains where it carries authorized weight) but fails to govern heart rate variability, cortisol baseline, or threat-detection thresholds (regulatory domains governed by $\mathcal{H}_{\text{auth}}$). Conversely, when governance shifts and $h_{\text{safe}}$ enters $\mathcal{H}_{\text{auth}}$, regulatory stabilization may occur rapidly, sometimes without observable gradual increases in precision or extended evidence accumulation, because the constraint on regulatory admissibility has changed, not the inferential strength of the hypothesis itself.This categorical distinction explains why interventions targeting precision (cognitive restructuring, evidence collection, exposure therapy) can modify explicit beliefs without producing durable autonomic stabilization, while governance-level shifts produce immediate cross-system coherence without requiring gradual precision increases.


ALPs vs. Context Priors
Context priors modulate which identity hypotheses are active based on environmental contingencies and are inferentially updated as contextual evidence accumulates (Pezzulo et al., 2018). A context prior might activate $h_{\text{threat}}$ in ambiguous social situations while activating $h_{\text{safe}}$ in familiar settings, with transitions driven by contextual cues and reversible through environmental change. ALPs, by contrast, define the regulatory regime determining which active hypotheses are permitted to regulate autonomic and behavioural systems, persisting across contexts without continuous environmental support. A context prior determines which hypothesis is contextually appropriate; an ALP determines which hypothesis---once active---is authorized to govern autonomic systems. The critical distinction emerges in cross-context persistence: context priors fluctuate with evidence and reverse when environmental signals shift, while ALP-constrained regulatory dominance maintains stability across diverse contexts without requiring sustained reinforcement. A human agent under governance-aligned regulation exhibits consistent autonomic baselines, affective set-points, and stress-recovery patterns whether in novel environments, relational challenges, or identity-threatening situations---stability that is not formally specified by contextual gating mechanisms alone.

ALPs vs. Inhibitory Control
Prefrontal inhibitory control suppresses responses generated by competing hypotheses through sustained effortful modulation, associated with increased engagement of prefrontal control networks to dampen limbic reactivity (Miller \& Cohen, 2001). This mechanism is metabolically expensive, depletes under cognitive load, and collapses when attentional resources are diverted. A human agent using inhibitory control to suppress anxiety generated by $h_{\text{threat}}$ must continuously monitor and override autonomic arousal, producing elevated dlPFC activity, increased subjective effort, and vulnerability to stress-induced relapse. ALP alignment, conversely, eliminates the need for suppression by authorizing hypotheses whose predictions are already aligned with regulatory goals, thereby reducing control demand rather than intensifying it. When $h_{\text{safe}} \in \mathcal{H}_{\text{auth}}$, autonomic predictions align with regulatory objectives rather than requiring prefrontal suppression of threat-based responses; the system generates safety-oriented predictions endogenously without requiring prefrontal override of autonomic threat responses. A predicted empirical distinction is that top-down inhibition is associated with increased prefrontal control engagement and sustained effort, while ALP alignment decreases prefrontal load and operates without compensatory control. Operationally, inhibition represents effortful suppression of unauthorized predictions; governance alignment represents structural permission eliminating prediction conflict at the regulatory interface.

ALPs vs. Metacognition
Metacognitive monitoring involves prefrontal assessment of lower-level cognitive states---detecting misalignment between conscious beliefs and autonomic predictions, evaluating prediction accuracy, and tracking confidence in one's own judgments (Fleming \& Dolan, 2012; Barrett \& Simmons, 2015). A human agent can metacognitively recognize the discrepancy between conscious endorsement of $h_{\text{safe}}$ and autonomic operation under $h_{\text{threat}}$, generating explicit awareness of the mismatch: ``I know intellectually that I am safe, yet my body responds as threatened.'' This detection, however, does not resolve the governance conflict. Metacognition monitors and evaluates; ALPs authorize and constrain. Metacognitive awareness can identify regulatory misalignment without possessing the architectural authority to shift which hypothesis regulates. The clinical phenomenon of ``insight without stability'' directly reflects this distinction---clients achieve metacognitive clarity about maladaptive patterns, accurately identify the mismatch between endorsed beliefs and autonomic predictions, yet regulatory dominance remains unchanged. ALPs operate at a level orthogonal to metacognitive monitoring: they determine which hypothesis is permitted to regulate, not which hypothesis is being monitored or evaluated. A governance shift may occur with or without metacognitive awareness, and metacognitive insight may be present with or without governance realignment.

Taken together, these distinctions position ALPs as meta-structural constraints on regulatory admissibility—categorically distinct from precision (scalar weighting), policy priors (preference ranking), Bayesian model selection (inferential comparison), context priors (reversible environmental modulation), inhibitory control (effortful suppression), and metacognition (monitoring and evaluation). The construct's necessity derives from the architectural requirement for stable regulatory coherence in systems maintaining multiple high-precision identity hypotheses, and its functional expression determines—orthogonal to inferential strength and contextual activation—which hypotheses interface with autonomic and behavioural control systems.

\subsection*{Glossary of Symbols and Notation}
\begin{table}[H]
\centering
\begin{tabular}{cl}
\hline
\textbf{Symbol} & \textbf{Definition} \\
\hline
$\mathcal{H}$ & Set of all coherent identity-level hypotheses \\
$\mathcal{H}_{\text{auth}}$ & Subset of hypotheses authorized for regulatory control \\
$h_i$ & Individual hypothesis (e.g., $h_{\text{safe}}$, $h_{\text{threat}}$) \\
$P(h_i | \text{data})$ & Posterior probability of hypothesis $h_i$ \\
$\pi_i$ & Precision (confidence weighting) of hypothesis $h_i$ \\
$C_{\text{auth}}$ & Constraint function determining regulatory admissibility \\
$\mathcal{P}$ & Set of all possible policies (action sequences) \\
$\pi^*$ & Optimal policy selected through expected free energy minimization \\
$G(\pi)$ & Expected free energy of policy $\pi$ \\
\hline
\end{tabular}
\caption{Key symbols and notation used throughout the paper.}
\label{tab:glossary}
\end{table}

\vspace{1cm}

\section{Computational Placement}

\subsection{ALPs and Precision-Weighting Architecture}

Within active inference frameworks, precision ($\pi$) functions as a gain parameter modulating the influence of prediction errors on belief updating and policy selection (Feldman \& Friston, 2010). High precision amplifies prediction error signals, driving stronger updates to posterior beliefs and increasing the likelihood that policies aligned with high-precision predictions will be selected. ALPs operate at a structurally distinct level from precision-weighting dynamics: they constrain regulatory admissibility as a boundary condition on which hypotheses can participate in precision-weighted competition within the inferential architecture.

Where precision modulates how strongly prediction errors influence inference, ALPs constrain which identity hypotheses are permitted to interface with regulatory control systems. Formally, precision operates within the likelihood function $P(\text{data} | h, \pi)$, scaling the weight of sensory evidence relative to prior expectations. ALPs, by contrast, define a constraint $C_{\text{auth}}: \mathcal{H} \to \{\text{authorized}, \text{unauthorized}\}$ that determines regulatory interface eligibility orthogonal to precision-weighting dynamics. Two hypotheses $h_i, h_j \in \mathcal{H}$ may carry identical precision $\pi_i = \pi_j$ yet differ categorically in regulatory authority: $C_{\text{auth}}(h_i) = \text{authorized}$ while $C_{\text{auth}}(h_j) = \text{unauthorized}$. This is not a scalar difference in gain; it is a categorical difference in control-system access. In this sense, ALPs function analogously to boundary-condition constraints in physical systems, constraining admissible trajectories without modifying the inferential calculus operating within them.

ALPs are therefore not reducible to hyperpriors over precision: hyperpriors modulate expected precision within the inferential hierarchy, whereas ALPs constrain the mapping from hypothesis to regulatory output independently of precision scaling.

\subsection{ALPs and Policy Space Constraints}

In active inference, agents minimize expected free energy $G(\pi)$ over policies $\pi \in \mathcal{P}$, where policies represent sequences of actions designed to reduce both uncertainty (epistemic value) and distance from preferred states (pragmatic value; Friston et al., 2017). 
\begin{align}
\pi^* &= \arg\min_{\pi \in \mathcal{P}} G(\pi)
\end{align}

Policy priors $P(\pi)$ encode preferences over action sequences, biasing selection toward certain behaviours. ALPs do not modify policy priors or expected free energy calculations directly; they constrain which identity-level hypotheses generate candidate policies for evaluation in the first place.
\begin{align}
\pi^* &= \arg\min_{\pi \in \mathcal{P}(h \in \mathcal{H}_{\text{auth}})} G(\pi)
\end{align}

The critical distinction: policy selection operates within an already-defined policy space $\mathcal{P}$; ALPs determine which identity hypotheses are permitted to generate policies that enter the regulatory-evaluable subset of the policy space, without modifying the expected free energy calculus itself. An unauthorized hypothesis $h_i \notin \mathcal{H}_{\text{auth}}$ may generate coherent, low-free-energy policies in representational domains (planning, explicit reasoning), yet those policies remain excluded from autonomic and affective regulatory control regardless of their computed value.

ALPs are not parameters within the generative model's likelihood structure ($A$ matrices) or state transition dynamics ($B$ matrices); they function as boundary conditions on the hypothesis-to-policy mapping within which expected free energy minimization operates. Policy selection chooses among evaluable actions; ALPs define which hypothesis-generated actions are evaluable at the regulatory level. This is a constraint on the hypothesis-to-policy mapping, not a change to the variational objective.

\subsection{ALPs and Prefrontal Arbitration Networks}

Prefrontal systems, particularly dorsolateral and ventromedial regions, are implicated in arbitration---resolving conflicts when multiple high-level representations compete for behavioral control (Miller \& Cohen, 2001; Domenech \& Koechlin, 2015). Anterior cingulate cortex detects prediction error and conflict, signaling when current predictions fail to minimize free energy (Shenhav et al., 2013). These systems operate within the standard predictive processing framework: they detect misalignment, modulate precision, and bias policy selection toward conflict-resolving actions. ALPs are distinct from these arbitration mechanisms. Arbitration resolves competition within the authorized hypothesis set $\mathcal{H}_{\text{auth}}$; ALPs specify the constraint structure that defines which hypotheses are eligible for regulatory arbitration. The anterior cingulate may detect conflict between $h_{\text{safe}}$ and $h_{\text{threat}}$, signalling elevated prediction error. Standard arbitration mechanisms then allocate control resources, potentially increasing precision on one hypothesis to resolve the conflict. However, if $h_{\text{safe}} \notin \mathcal{H}_{\text{auth}}$, arbitration mechanisms cannot assign it regulatory control if it remains outside the admissible hypothesis set, regardless of precision adjustments within representational domains. Detection of regulatory misalignment (a metacognitive and conflict-monitoring function) does not confer authorization to regulate. ALPs specify the constraint structure that defines which hypotheses are eligible candidates for autonomic and behavioral control through arbitration processes.

\subsection{Functional Neurobiological Expression}

ALPs are not localized to a single neural substrate but are functionally expressed through distributed control networks involved in rule maintenance, policy selection, and top-down modulation. Dorsolateral prefrontal cortex supports the maintenance of abstract rules and task-set representations that gate lower-level processing (Miller \& Cohen, 2001). Ventromedial prefrontal cortex integrates affective and visceral predictions with goal representations, influencing autonomic set-points and policy valuation (Roy et al., 2012). Anterior cingulate cortex monitors conflict and allocates control resources (Shenhav et al., 2013). These systems collectively implement the architecture through which governance constraints are expressed---not as localized ALP modules, but as emergent properties of distributed control dynamics. If ALPs are valid, governance-stable states would be expected to correlate with coordinated activity across these networks, including reduced conflict signaling (ACC), decreased compensatory control engagement (dlPFC), and coherent autonomic prediction integration (vmPFC). Critically, ALPs are defined functionally through system-level regulatory behavior, not anatomically through regional activation. Neurobiological correlates provide empirical markers of governance states but do not constitute the constraint mechanism itself, which operates as an architectural property of the control hierarchy.

\subsection{Dynamical Implications and Parsimony}

From a dynamical systems perspective, ALPs function as constraints on accessible attractor basins within the hypothesis-policy landscape. Governance shifts do not represent transitions between existing attractors (which would be explained by standard noise perturbations or precision fluctuations); they represent changes in state-space boundary conditions that determine basin accessibility---a constraint on which attractors are evaluable, not the creation of new dynamical variables.

This should be understood as a change in boundary conditions governing state-space accessibility, not as the creation of new dynamical variables or violation of variational principles. A hypothesis $h_i$ that was previously ineligible for regulatory control (operating in a representational attractor with no pathway to autonomic regulation) becomes eligible when $C_{\text{auth}}(h_i)$ shifts, opening previously inaccessible regions of the control landscape. This accounts for rapid, durable stabilization without gradual evidence accumulation: the system transitions into a newly accessible basin rather than slowly climbing out of an existing one.

Introducing ALPs does not violate parsimony. The construct does not introduce additional representational content or new inferential operators; it formalizes a constraint relation that is already implicitly required for regulatory coherence in systems maintaining multiple high-precision identity hypotheses---namely, the specification of which hypotheses are permitted to interface with autonomic and behavioral control systems. Without this formalization, the architecture may remain under-specified at the governance level, unable to account for regulatory dominance persistence, cross-context stability, or the categorical distinction between representational and regulatory authority. ALPs complete the architecture by defining the constraint structure that precision, policy selection, and arbitration operate within.

\begin{figure}[H]
\centering
\includegraphics[width=0.75\textwidth]{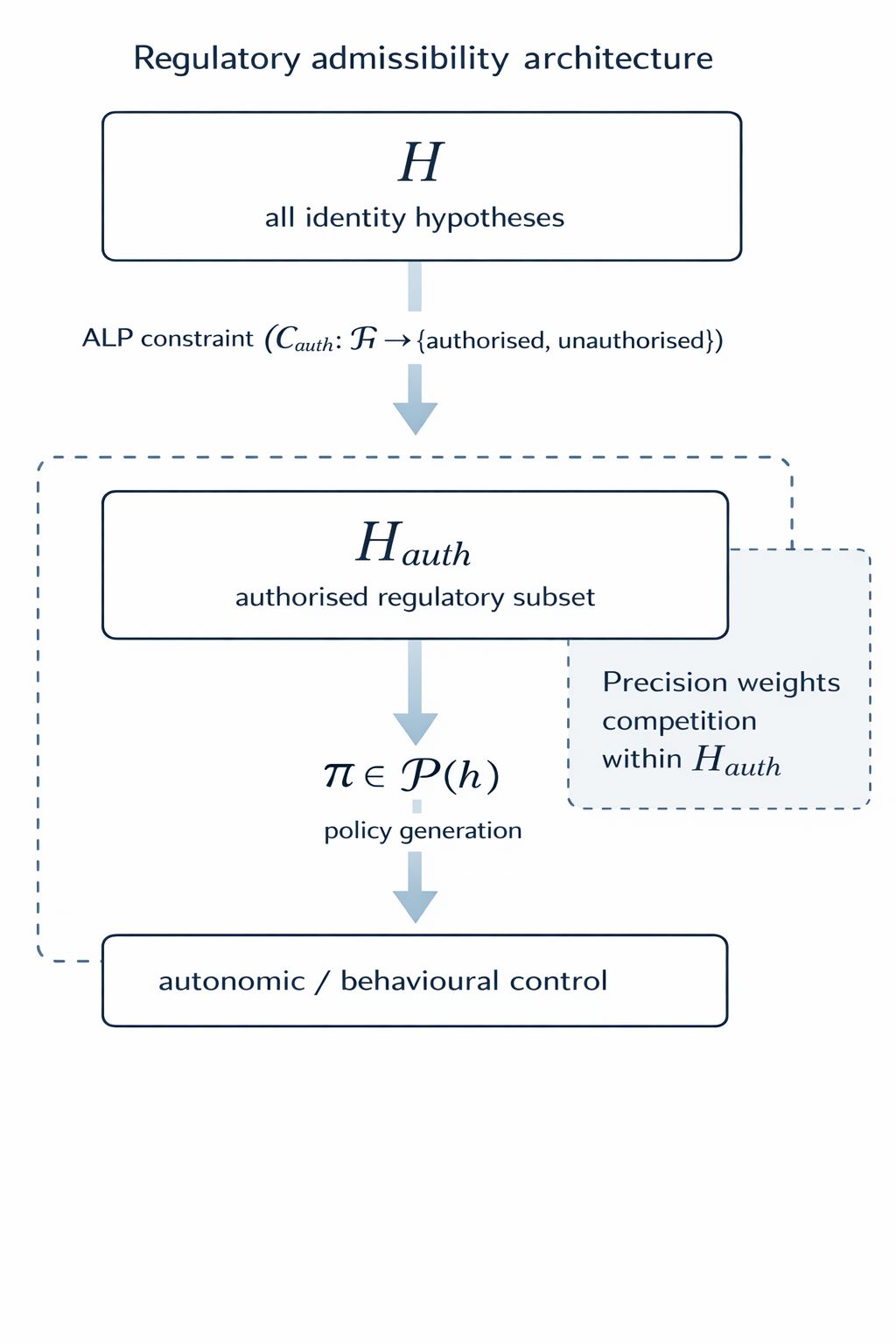}
\caption{Regulatory admissibility architecture. Authority-Level Priors constrain the subset of identity hypotheses permitted to generate policies that interface with autonomic and behavioural regulation. Precision weighting operates only within this authorised subset. This distinction formalises why representational belief updating can occur without corresponding autonomic stabilisation when an adaptive hypothesis remains outside $\mathcal{H}_{\text{auth}}$.}
\label{fig:alp_architecture}
\end{figure}




\section{Explaining Durable vs.\ Non-Durable Change}

Operational Criteria for Durable Change
Durable change in the present framework is defined through measurable regulatory stability, not subjective coherence or conscious conviction. A shift qualifies as durable when: (1) autonomic baselines stabilize across contexts without continuous environmental reinforcement, (2) prediction-error volatility decreases under identity-relevant stressors, (3) compensatory prefrontal control engagement (dlPFC activity) reduces during regulatory challenges, and (4) relapse probability under stress induction diminishes relative to baseline. Non-durable change, by contrast, exhibits context-dependent stability that collapses when environmental support withdraws, elevated conflict-monitoring activity (ACC) persisting despite conscious belief updates, sustained requirement for effortful cognitive control to maintain regulatory coherence, and high vulnerability to stress-induced return to prior regulatory states. These criteria distinguish structural regulatory reassignment from temporary precision modulation or context-dependent belief updates that fail to persist under load.

\subsection{Insight Without Stability: 
The Representational-Regulatory Divide}

A recurrent clinical phenomenon challenges standard predictive processing accounts: clients achieve sustained insight into maladaptive identity patterns, consciously endorse alternative self-models with high conviction, accumulate substantial therapeutic evidence supporting new beliefs, yet autonomic threat responses and stress-reactivity patterns remain unchanged. This is not occasional therapeutic resistance but a recurrent asymmetry observed across multiple intervention modalities (Hayes et al., 2011; Bouton, 2002). Standard explanations invoke context-dependent precision reweighting---therapeutic contexts assign high precision to adaptive beliefs while stressful contexts reinstate precision on threat-based priors---or incomplete memory reconsolidation, where original learning remains intact despite new belief formation.

These mechanisms capture important dynamics but do not formally specify why regulatory dominance persists even when contextual evidence and precision adjustments consistently favor alternative hypotheses across multiple domains and extended timeframes. Specifically, these accounts explain why a previously reinforced hypothesis may regain precision under stress, but they do not specify why an extensively reinforced alternative hypothesis fails to acquire regulatory authority despite sustained posterior dominance across contexts.

Consider a client who, through sustained cognitive restructuring, increases precision on $h_{\text{competent}}$, reducing uncertainty and accumulating confirming evidence within representational and explicit reasoning domains across therapeutic, professional, and relational contexts. Posterior probability $P(h_{\text{competent}} \mid \text{data})$ increases substantially; explicit endorsement strengthens; metacognitive awareness of the shift is present. Yet under moderate stress---a challenging work presentation, unexpected criticism, relational ambiguity---autonomic regulation reverts to patterns consistent with $h_{\text{inadequate}}$: stress-related autonomic activation, hypervigilance, defensive policy selection, and threat-oriented affective predictions. If precision weighting alone determined regulatory influence, the extensively reinforced, high-precision $h_{\text{competent}}$ hypothesis should dominate. Its persistent failure to govern autonomic and affective systems despite favorable posterior probability is consistent with the hypothesis remaining excluded from regulatory authorization despite acquiring high inferential strength.
\vspace{1cm}
The ALP framework proposes that $h_{\text{competent}} \notin \mathcal{H}_{\text{auth}}$---the hypothesis remains excluded from regulatory authorization despite acquiring high inferential strength. Therapeutic work successfully updates representational content and increases precision within that domain, yet regulatory admissibility remains unchanged because the governance constraint determining which hypothesis regulates under stress remains structurally fixed. Standard predictive processing accounts do not reject these mechanisms (context-dependent retrieval, precision modulation, memory reconsolidation) but argue they are insufficient to explain why regulatory dominance persists when both contextual evidence and sustained precision strengthening favor alternative hypotheses. Even when posterior probability and explicit endorsement converge toward an alternative identity hypothesis, regulatory authority may remain assigned to a legacy hypothesis, indicating that inferential updating alone does not guarantee regulatory reassignment. This persistence cannot be reduced to insufficient precision, as precision on the alternative hypothesis may exceed that of the legacy hypothesis within representational domains while remaining excluded from regulatory interface. This asymmetry---representational belief updated, regulatory control unchanged---constitutes the empirical signature that ALPs are proposed to formalize.

\subsection{Potential Clinical Implications and Misclassification Risk}

An additional implication concerns the possibility of misclassification during transitional regulatory states. If governance-level reassignment is occurring or destabilizing, yet is interpreted solely as insufficient belief conviction or inadequate expectation strengthening, interventions may intensify precision-targeted strategies (e.g., repeated cognitive restructuring, increased exposure demands) without addressing regulatory admissibility. Within the present framework, such misalignment could theoretically reinforce effortful inhibitory control rather than facilitate structural reassignment, potentially prolonging instability under stress. This distinction is critical: placebo and expectation effects operate through precision modulation and belief strengthening, whereas governance reassignment involves structural changes in regulatory admissibility that are orthogonal to conscious conviction or expectation strength. Similar considerations may apply in contexts involving transient reductions in hierarchical precision, such as psychedelic-assisted interventions, acute stress responses, or meditation-induced absorption states, where expanded representational flexibility is often interpreted as therapeutic reorganisation. Even if entropy increase permits representational reorganisation, regulatory admissibility constitutes a separate structural layer. Durable change would therefore require reassignment within the regulatory-admissible set rather than precision relaxation alone. This does not imply that precision-based or entropy-based mechanisms are ineffective; rather, it highlights the importance of distinguishing representational updating from regulatory authorization when evaluating treatment response. Empirical investigation would be required to determine whether identifiable transitional markers precede durable governance shifts and how intervention timing may interact with these dynamics.

\subsection{Repetition Failure Under Stress: Neuromodulation vs.\ Governance}

Therapeutic interventions based on repetition---exposure therapy, behavioral activation, skill rehearsal---often produce context-dependent improvements that collapse under acute stress. Standard accounts attribute this to neuromodulatory shifts: stress elevates noradrenaline and cortisol, which modulate precision weighting by increasing gain on threat-related prediction errors and decreasing gain on safety signals (Hermans et al., 2014). This explanation captures how stress amplifies certain priors but does not specify why the same identity hypothesis repeatedly regains regulatory dominance across diverse stressors and contexts, even when posterior probability and representational precision favour an alternative hypothesis in non-stressed states. Neuromodulation explains the mechanism of gain adjustment; it does not explain the structural constraint determining which hypothesis receives amplified gain under stress. The ALP framework proposes that stress-induced relapse reflects the persistence of $h_{\text{threat}} \in \mathcal{H}_{\text{auth}}$: when neuromodulatory systems increase overall arousal, regulatory control defaults preferentially to the authorized hypothesis, even when alternative hypotheses have acquired higher representational precision through therapeutic work. Neuromodulation modulates gain; ALPs constrain which hypotheses that gain is permitted to amplify at the regulatory interface.

\subsection{Optimization Plateau: Residual Regulatory Constraint}

Performance optimization across domains---cognitive enhancement, athletic training, therapeutic progress---frequently exhibits a plateau where further incremental effort produces diminishing returns despite continued practice and refined technique. This pattern can be framed computationally: improvements plateau when regulatory authority remains assigned to threat-based or self-protective identity hypotheses, generating residual prediction error under performance-relevant stress that cannot be resolved through incremental precision adjustments alone. A highly trained individual may possess extensive skill-based predictions ($h_{\text{skilled}}$) with low uncertainty and high posterior probability, yet if $h_{\text{inadequate}} \in \mathcal{H}_{\text{auth}}$, performance under evaluative pressure will reflect the regulatory hypothesis rather than the representational one. The system minimizes prediction error within the constraint structure imposed by authorized hypotheses; when that structure includes threat-based predictions at the regulatory level, further skill acquisition reduces error in execution domains but cannot eliminate the regulatory prediction error generated by identity-level threat models. The plateau is consistent with an architectural ceiling imposed by governance constraints, not a limit in skill acquisition or precision refinement. Standard optimization approaches target precision and policy improvement within the existing regulatory framework; they do not address the governance-level constraint determining which identity hypothesis regulates under pressure.

\subsection{Rapid Durable Stabilization: Boundary Condition Reconfiguration}

Clinical observations document instances where regulatory patterns shift rapidly and persist across contexts without requiring extended evidence accumulation or gradual precision increases (sudden gains literature: Tang \& DeRubeis, 1999). Standard predictive processing accounts struggle to explain such shifts without invoking catastrophic model replacement or extreme precision spikes, both computationally demanding and phenomenologically inconsistent with reported experiences. The ALP framework proposes that rapid durable stabilization reflects governance-level reassignment: $h_{\text{safe}}$ transitions into $\mathcal{H}_{\text{auth}}$ through boundary condition reconfiguration rather than through incremental belief revision or precision accumulation. This does not imply discontinuous Bayesian updating of hypothesis content; rather, it reflects a shift in regulatory admissibility within which updating proceeds. Critically, the representational content of both $h_{\text{safe}}$ and $h_{\text{threat}}$ may remain largely unchanged---what shifts is regulatory permission, not belief structure. This explains why such transitions can occur rapidly (no gradual evidence accumulation required), produce cross-context stability (regulatory authorization persists independently of contextual fluctuations), and reduce compensatory control demands (hypotheses authorized at the regulatory interface generate coherent regulatory predictions without prefrontal override). These patterns are consistent with governance reassignment rather than precision-based belief updating, though the mechanisms triggering such reassignments remain empirically under-specified and constitute a priority for future investigation.

\section{Falsifiable Markers and Empirical Predictions}

Inferring ALPs as Latent Constraints
Authority-Level Priors are not directly observable variables but are proposed as latent constraints inferred from patterns of regulatory behaviour that standard precision-based models do not parsimoniously account for. This is not reverse inference from isolated behavioural changes but the identification of a structural variable that explains a specific covariance signature—the joint occurrence of rapid stabilisation, cross-context persistence, reduced compensatory control, and stress resilience without extended rehearsal. Individual markers (reduced autonomic reactivity, improved affect regulation, decreased prefrontal engagement) can result from multiple mechanisms: exposure therapy, pharmacological intervention, habit formation, or neuromodulatory shifts. What distinguishes governance reassignment from these alternatives is the simultaneous emergence of (1) reduced stress reactivity under identity-relevant challenges, (2) decreased compensatory prefrontal control engagement without performance degradation, (3) cross-context regulatory stability without context-specific training, and (4) resilience to stress reinstatement without requiring sustained environmental support or cognitive rehearsal. This covariance pattern is not straightforwardly explained by precision modulation alone, which typically predicts improvements proportional to exposure history and context-specific evidence accumulation. ALPs are proposed as the most parsimonious latent constraint accounting for this joint profile. The ALP construct would be weakened if these markers dissociate - for example, if reduced stress reactivity consistently occurs without reductions in compensatory control or without cross-context persistence.

\vspace{1cm}
Stress Reactivity Curves: The ALP Signature
Standard interventions targeting precision (cognitive restructuring, exposure therapy) produce stress-reactivity improvements that follow characteristic patterns: gradual reduction in peak autonomic arousal across repeated exposures, context-dependent stability requiring rehearsal in target environments, and vulnerability to relapse when novel stressors are introduced without prior exposure gradients. Governance reassignment predicts a distinct signature consistent with altered regulatory admissibility rather than incremental precision accumulation: rapid reduction (within a limited number of identity-relevant exposures, without graded exposure progression) in autonomic reactivity across diverse identity-relevant stressors (not limited to trained contexts), faster return-to-baseline following stress induction without requiring compensatory cognitive control (measured via heart rate variability recovery time constants and skin conductance decline slopes), and maintained stability when novel identity-threatening situations are introduced without prior contextual exposure. Critically, this stability emerges without elevated dorsolateral prefrontal cortex activity - indicating regulatory coherence rather than effortful suppression. The distinguishing pattern is not lower reactivity alone (which could reflect habituation or avoidance) but the combination of: reduced peak autonomic response without concomitant reductions in task engagement or behavioral performance, accelerated recovery dynamics, diminished prefrontal compensatory engagement, and generalization to untrained contexts. This profile is inconsistent with precision-based gradual learning, which requires context-specific evidence accumulation and produces stability proportional to exposure frequency.

\subsection{Computational Formalization and Predictions}

The ALP constraint can be formalized minimally within active inference. 

Standard policy selection minimizes expected free energy over all available policies:
\begin{equation}
\pi^* = \arg\min_{\pi \in \mathcal{P}} G(\pi)
\label{eq:standard_policy}
\end{equation}
\vspace{1cm}

Under ALP constraints, policy selection operates only over policies generated by authorized hypotheses:
\begin{equation}
\pi^* = \arg\min_{\pi \in \mathcal{P}(h \in \mathcal{H}_{\text{auth}})} G(\pi)
\label{eq:alp_policy}
\end{equation}
\vspace{1cm}

This constraint is not equivalent to adjusting policy priors $P(\pi)$, as those priors operate within the admissible set, whereas ALPs constrain the admissible set itself. This formalization generates a testable prediction: if $h_i \notin \mathcal{H}_{\text{auth}}$, increasing posterior probability $P(h_i \mid \text{data})$ or precision $\pi_i$ on that hypothesis will not necessarily alter $\pi^*$ under identity-relevant stress, because policies generated by $h_i$ remain excluded from the evaluable set regardless of inferential strength, as admissibility precedes weighting within the policy-selection process. Empirically, this predicts that therapeutic interventions successfully increasing precision on adaptive identity hypotheses (measurable via reduced uncertainty in explicit self-reports and stronger conviction ratings) may fail to produce stress-reactivity changes when those hypotheses remain outside $\mathcal{H}_{\text{auth}}$. Conversely, governance shifts should produce regulatory changes without requiring proportional increases in precision or posterior probability.
\vspace{1cm}

Experimental Paradigms and Empirical Markers
Testing ALP predictions requires paradigms that dissociate representational updating from regulatory control. One approach: measure trial-to-trial variance in error-related negativity (ERN) and anterior cingulate BOLD signal fluctuation during identity-relevant performance tasks (Trier Social Stress Test, evaluative public speaking) before and after interventions. Governance-level reassignment predicts reduced variance in conflict-monitoring signals across repeated stress exposures without requiring escalating compensatory control (stable low dlPFC engagement), whereas precision-based improvement predicts either sustained high dlPFC activity (effortful control) or gradual reduction tied to exposure frequency. Autonomic markers include: heart rate variability recovery time constants (time to return to baseline following stress offset), skin conductance response habituation slopes across novel identity-threatening scenarios, and cortisol response curves under evaluative challenge. Cross-context generalization can be tested by introducing novel identity-relevant stressors without prior exposure or rehearsal, measuring whether regulatory stability transfers. Critically, performance on identity-relevant tasks (accuracy, response times, error rates) must remain stable or improve alongside reduced prefrontal engagement - distinguishing regulatory efficiency from disengagement or avoidance

\vspace{1cm}
Falsification Criteria
The ALP framework generates clear falsification conditions. The construct would be disconfirmed if: (1) increasing posterior probability and precision on alternative identity hypotheses reliably produces cross-context regulatory reassignment under stress without increased compensatory control load, indicating precision alone is sufficient; (2) the proposed covariance signature consistently dissociates, with markers occurring independently rather than jointly (e.g., reduced reactivity without cross-context persistence, or rapid change without durability); (3) stress-reactivity changes consistently require proportional increases in prefrontal compensatory control, indicating all regulatory shifts operate through effortful inhibition rather than permission structure changes; or (4) no measurable differences emerge between interventions targeting precision (standard CBT, exposure) and those hypothesized to engage governance mechanisms, when controlling for exposure frequency and therapeutic alliance. A comprehensive test of the framework would require longitudinal designs incorporating stress induction, computational modelling of policy selection, and neurophysiological measurement of conflict-monitoring and control dynamics. These criteria provide empirical leverage: ALPs are not unfalsifiable architectural assumptions but testable latent constraints governing admissibility, inferred from specific measurable regulatory patterns that should dissociate from precision-based mechanisms if the framework is valid.

\section{Discussion and Implications}

Theoretical Implications
The present framework proposes that hierarchical predictive processing architectures require a formally specified governance-level constraint to account for regulatory stability in systems with identity-level self-models. Authority-Level Priors are not introduced as a replacement for precision-weighting mechanisms, policy priors, or contextual modulation - all of which remain essential components of predictive processing - but as the formalization of a constraint structure that these mechanisms operate within. The framework clarifies the distinction between representational updating (belief revision, posterior probability shifts, explicit endorsement) and regulatory reassignment (authorization to interface with autonomic and behavioral control systems), offering a potential explanation for why therapeutic interventions can successfully modify conscious beliefs and increase precision on adaptive hypotheses without producing corresponding changes in stress reactivity, affective regulation, or behavioral policy selection under identity-relevant challenge. This formalizes a governance-level constraint that hierarchical architectures appear to require for modelling identity-level regulation. The proposal is compatible with existing formulations of active inference and may be incorporated through explicit modelling of admissible hypothesis sets without altering the underlying variational framework.\vspace{1cm}

Clinical Implications
The framework may provide an additional lens for understanding why some interventions yield durable regulatory change while others remain context-bound despite comparable therapeutic alliance, exposure frequency, and explicit belief modification. Standard clinical measures focus predominantly on representational change - symptom self-report, explicit belief ratings, conscious insight - yet these may dissociate from regulatory stabilization if governance constraints remain unchanged. The ALP framework suggests that measuring regulatory markers (autonomic recovery dynamics, compensatory control engagement, cross-context stress resilience) alongside representational markers could refine intervention targeting and outcome assessment. This is not a prescription for specific therapeutic techniques but an architectural consideration: interventions targeting precision (evidence accumulation, cognitive restructuring) may succeed at representational updating while failing to engage governance-level reassignment, potentially contributing to explanations of why insight-rich therapeutic progress sometimes fails to translate to stress-resilient behavioral change. The framework invites investigation into whether certain intervention modalities differentially engage governance mechanisms versus precision modulation, though such differentiation remains empirically under-specified and constitutes a priority for controlled comparison studies.

\vspace{1cm}
Optimization and Performance Plateaus
Performance optimization across domains frequently encounters plateaus where incremental effort yields diminishing returns. The ALP framework may contribute to understanding this phenomenon computationally: regulatory constraints may limit further prediction-error reduction under evaluative stress despite continued skill acquisition and precision refinement. When regulatory authority remains assigned to threat-based or self-protective identity hypotheses, skill-based predictions may exhibit high accuracy and low uncertainty in practice contexts, yet performance under evaluative pressure reflects the regulatory hypothesis rather than the skill-based representational hypothesis. This suggests a distinction between execution-level optimization (skill acquisition, policy refinement) and regulatory-level constraints (governance over stress-responsive systems), though empirical validation of this distinction requires controlled studies dissociating skill measures from regulatory markers under stress induction.
\vspace{1cm}
Rapid Regulatory Change and Spontaneous Phenomena
The framework may also clarify distinctions between governance-level reassignment and expectation-based mechanisms such as placebo effects. Placebo responses are typically modelled within predictive processing as modulation of prior precision and outcome expectancy—altering prediction-error weighting through belief about likely improvement. By contrast, ALPs are not expectation priors and are not defined by representational confidence regarding anticipated outcomes. They operate at the level of regulatory admissibility, constraining which identity-level hypotheses are permitted to interface with autonomic and behavioural control systems independently of posterior probability or expected value. If certain rapid regulatory shifts reflect governance-level reassignment rather than precision modulation, they would be expected to exhibit the predicted covariance signature: rapid onset, cross-context persistence, reduced compensatory control engagement, and stress resilience without requiring elevated expectancy or sustained belief reinforcement. This distinction is critical: expectation effects operate through inferential weighting within the authorised hypothesis set, whereas governance shifts alter the authorised set itself. Empirical paradigms would therefore need to dissociate expectancy-driven precision effects from structural changes in regulatory admissibility using controlled stress-induction and longitudinal physiological measurement.

\subsection{Limitations and Future Directions}
Several limitations and future directions warrant clarification. First, the mechanisms triggering governance-level reassignments remain empirically unspecified---what conditions enable $h_i$ to transition into or out of $\mathcal{H}_{\text{auth}}$ constitutes a central target for future investigation. Second, developmental origins of governance constraints are not addressed; whether such constraints emerge through attachment dynamics, evolutionary pressures, cultural transmission, or other mechanisms remains an open empirical question. Third, neurobiological mapping remains provisional and functional rather than mechanistic---distributed control network activity provides plausible correlates of governance states but does not yet establish implementation-level specification. Fourth, the computational formalisation presented is intentionally minimal and requires expansion through explicit generative modelling, agent-based simulations, and formal derivations to determine whether governance constraints can be derived from, or must be added to, existing hierarchical frameworks. Finally, alternative explanations---including hierarchical precision restructuring, hyperprior dynamics, or nonlinear attractor reconfiguration---remain empirically viable until discriminative experimental evidence is obtained. The present proposal should therefore be understood not as a completed explanatory model, but as a formally articulated architectural hypothesis that introduces a tractable target for computational and longitudinal validation.
\vspace{1cm}

Some interpretive frameworks describe governance-level constraints in relational or philosophical terms. The present account remains agnostic regarding ultimate origin and restricts its claims to functional and computational architecture.

\section*{Acknowledgments}

The author declares no conflicts of interest. No external funding was received for this work.

\section*{Data Availability Statement}

This is a theoretical paper. No empirical data were collected or analysed.

\section*{References}
\setlength{\parindent}{0pt}
\setlength{\parskip}{0.5em}

Barrett, L. F. (2017). \textit{How emotions are made: The secret life of the brain}. Houghton Mifflin Harcourt.

Barrett, L. F., \& Simmons, W. K. (2015). Interoceptive predictions in the brain. \textit{Nature Reviews Neuroscience}, \textit{16}(7), 419--429. \url{https://doi.org/10.1038/nrn3950}

Bouton, M. E. (2002). Context, ambiguity, and unlearning: Sources of relapse after behavioral extinction. \textit{Biological Psychiatry}, \textit{52}(10), 976--986. \url{https://doi.org/10.1016/S0006-3223(02)01546-9}

Clark, A. (2016). \textit{Surfing uncertainty: Prediction, action, and the embodied mind}. Oxford University Press.

Domenech, P., \& Koechlin, E. (2015). Executive control and decision-making in the prefrontal cortex. \textit{Current Opinion in Behavioral Sciences}, \textit{1}, 101--106. \url{https://doi.org/10.1016/j.cobeha.2014.10.007}

Feldman, H., \& Friston, K. J. (2010). Attention, uncertainty, and free-energy. \textit{Frontiers in Human Neuroscience}, \textit{4}, Article 215. \url{https://doi.org/10.3389/fnhum.2010.00215}

Fleming, S. M., \& Dolan, R. J. (2012). The neural basis of metacognitive ability. \textit{Philosophical Transactions of the Royal Society B: Biological Sciences}, \textit{367}(1594), 1338--1349. \url{https://doi.org/10.1098/rstb.2011.0417}

Friston, K. (2010). The free-energy principle: A unified brain theory? \textit{Nature Reviews Neuroscience}, \textit{11}(2), 127--138. \url{https://doi.org/10.1038/nrn2787}

Friston, K., FitzGerald, T., Rigoli, F., Schwartenbeck, P., \& Pezzulo, G. (2017). Active inference: A process theory. \textit{Neural Computation}, \textit{29}(1), 1--49. \url{https://doi.org/10.1162/NECO_a_00912}

Hayes, S. C., Villatte, M., Levin, M., \& Hildebrandt, M. (2011). Open, aware, and active: Contextual approaches as an emerging trend in the behavioral and cognitive therapies. \textit{Annual Review of Clinical Psychology}, \textit{7}, 141--168. \url{https://doi.org/10.1146/annurev-clinpsy-032210-104449}

Hermans, E. J., Henckens, M. J. A. G., Joëls, M., \& Fernández, G. (2014). Dynamic adaptation of large-scale brain networks in response to acute stressors. \textit{Trends in Neurosciences}, \textit{37}(6), 304--314. \url{https://doi.org/10.1016/j.tins.2014.03.006}

Hohwy, J. (2013). \textit{The predictive mind}. Oxford University Press.

Miller, E. K., \& Cohen, J. D. (2001). An integrative theory of prefrontal cortex function. \textit{Annual Review of Neuroscience}, \textit{24}, 167--202. \url{https://doi.org/10.1146/annurev.neuro.24.1.167}

Parr, T., \& Friston, K. J. (2019). Generalised free energy and active inference. \textit{Biological Cybernetics}, \textit{113}(5--6), 495--513. \url{https://doi.org/10.1007/s00422-019-00805-w}

Pezzulo, G., Rigoli, F., \& Friston, K. J. (2018). Hierarchical active inference: A theory of motivated control. \textit{Trends in Cognitive Sciences}, \textit{22}(4), 294--306. \url{https://doi.org/10.1016/j.tics.2018.01.009}

Roy, M., Shohamy, D., \& Wager, T. D. (2012). Ventromedial prefrontal-subcortical systems and the generation of affective meaning. \textit{Trends in Cognitive Sciences}, \textit{16}(3), 147--156. \url{https://doi.org/10.1016/j.tics.2012.01.005}

Seth, A. K., \& Friston, K. J. (2016). Active interoceptive inference and the emotional brain. \textit{Philosophical Transactions of the Royal Society B: Biological Sciences}, \textit{371}(1708), 20160007. \url{https://doi.org/10.1098/rstb.2016.0007}

Shenhav, A., Botvinick, M. M., \& Cohen, J. D. (2013). The expected value of control: An integrative theory of anterior cingulate cortex function. \textit{Neuron}, \textit{79}(2), 217--240. \url{https://doi.org/10.1016/j.neuron.2013.07.007}

Tang, T. Z., \& DeRubeis, R. J. (1999). Sudden gains and critical sessions in cognitive-behavioral therapy for depression. \textit{Journal of Consulting and Clinical Psychology}, \textit{67}(6), 894--904. \url{https://doi.org/10.1037/0022-006X.67.6.894}

Wilkinson-Ryan, T., \& Westen, D. (2000). Identity disturbance in borderline personality disorder: An empirical investigation. \textit{American Journal of Psychiatry}, \textit{157}(4), 528--541. \url{https://doi.org/10.1176/appi.ajp.157.4.528}

\end{document}